\documentclass[10pt,twocolumn,letterpaper]{article}

\usepackage{iccv}
\usepackage{times}
\usepackage{epsfig}
\usepackage{graphicx}
\usepackage{amsmath}
\usepackage{amssymb}
\usepackage{enumitem}

\iccvfinalcopy 

\begin{document}

\title{Sparse to Dense Motion Transfer for Face Image Animation}

\author{Ruiqi Zhao$^{1,2}$,  Tianyi Wu$^{1,2}$,  Guodong Guo$^{1,2}$ \\
$^1$Institute of Deep Learning, Baidu Research, Beijing, China\\
$^2$National Engineering Laboratory for Deep Learning Technology and Application,
Beijing, China \\
{\tt\small\{zhaoruiqi, wutianyi01, guoguodong01\}@baidu.com}
}

\maketitle

\begin{abstract}
Face image animation from a single image has achieved remarkable progress. However, it remains challenging when only sparse landmarks are available as the driving signal. Given a source face image and a sequence of sparse face landmarks, our goal is to generate a video of the face imitating the motion of landmarks. We develop an efficient and effective method for motion transfer from sparse landmarks to the face image. We then combine global and local motion estimation in a unified model to faithfully transfer the motion. The model can learn to segment the moving foreground from the background and generate not only global motion,  such as rotation and translation of the face,  but also subtle local motion such as the gaze change. We further improve face landmark detection on videos. With temporally better aligned landmark sequences for training, our method can generate temporally coherent videos with higher visual quality. Experiments suggest we achieve results comparable to the state-of-the-art image driven method on the same identity testing and better results on cross identity testing.
\end{abstract}

\section{Introduction}


We consider the task of face image animation from a single image. Given a source image of a face and a sequence of face landmarks, our goal is to generate a video of the face imitating the motion of landmarks. It has a wide range of applications in video games, filming industry, retails, news broadcasting and teleconferencing among others. It has been a long standing problem of interest in computer graphics. Many works require 3D modelling and a large amount of training data for a specific person \cite{thies2016face2face, suwajanakorn2017synthesizing, thies2018headon}. Though capable of generating high quality videos, they have difficulty in generalizing to unseen identities. With the advancement of deep learning techniques, quite a few works \cite{wiles2018x2face, pumarola2018ganimation, zakharov2019few, Siarohin_2019_NeurIPS} are proposed for face image animation when only a few images or even one image of the face is given. Meanwhile, several face video datasets such as VoxCeleb \cite{Nagrani19, Chung18b,Nagrani17} and FaceForensics \cite{roessler2019faceforensicspp} are collected that facilitate the development of this challenging task.

Recently, Siarohin et al. \cite{Siarohin_2019_NeurIPS} develops a two-step approach for object-agnostic image animation. It first employs first-order motion model for dense motion estimation and then performs image refinement using an image generator. It is trained in a self-supervised manner and significantly improves the quality of animated face videos. The identity of source face is well maintained and the face motion has lots of details and is temporally coherent. Inspired by the success of \cite{Siarohin_2019_NeurIPS}, we also use the dense motion estimation followed by image generation approach to develop our method. Since \cite{Siarohin_2019_NeurIPS} assumes local affine motion, affine transformation matrices around a set of keypoints need to be extracted from source/driving image pairs. Different from them, we do not make assumptions on the motion model and only use sparse face landmarks as the driving signal. Our face landmarks are pre-defined and represent rich semantic meaning of the face. We directly conduct motion transfer from sparse face landmarks to the image.

We propose a simple method to transfer motion from sparse face landmarks to the face image. Specifically, we use adaptive instance normalization (AdaIN) layer \cite{karras2019style, huang2017arbitrary} and our Add\_Motion layer as the building blocks to fulfill this goal. AdaIN is invented for image style transfer and has been applied for face image animation \cite{zakharov2019few}. In \cite{zakharov2019few}, landmarks are rasterized to an image and AdaIN is used to transfer face identity. Different from them, we concatenate the coordinates of the landmarks and form a low-dimensional vector. We attempt to transfer geometry and motion. 

Eyes convey important and engaging message in communication. Therefore it is important to generate gaze change for realistic face image animation. Many existing works do not take it into account in their models \cite{zakharov2019few, wang2019few, zakharov2020fast}. We find it hard to achieve this goal when we apply a single motion transfer network on the whole image. To tackle this issue, we combine a global branch and three local branches in one unified network for motion generation. The global branch looks at the whole image and the local branches each only attends to the left eye, the right eye and the mouth region. The outputs of the four branches are combined to get the final dense motion map.

Being able to generate temporally coherent videos with high visual quality is challenging and it determines the applicability of the derived algorithm. To our best knowledge, existing works that only use landmarks as the driving signal do not consider the influence of detected landmarks quality \cite{zakharov2019few, wang2019few, zakharov2020fast}. We improve face landmark detection on videos by combining a heatmap prediction network \cite{bulat2017far} with a differentiable regression layer \cite{nibali2018numerical} as well as augmenting the regression loss with registration loss (SBR) \cite{dong2018supervision, dong2020supervision}. We additionally add eye pupils detection to help control gaze.

The contributions of our work are:
\begin{itemize}[noitemsep,nolistsep]
\item We propose an Add\_Motion layer, together with our novel application of AdaIN layer, we are able to transfer motion from sparse landmarks to face images.
\item Our global and local motion generation approach allows us to capture global face rotation and translation as well as subtle local motion such as the gaze change.
\item  We improve face landmark detection on videos and can generate videos with higher visual quality and better temporal coherence. 
\item Our method achieves comparable results as the state-of-the-art image driven methods on the same identity testing, and better results on cross identity testing. 
\end{itemize}

\begin{figure*}
\begin{center}
\includegraphics[width=\linewidth]{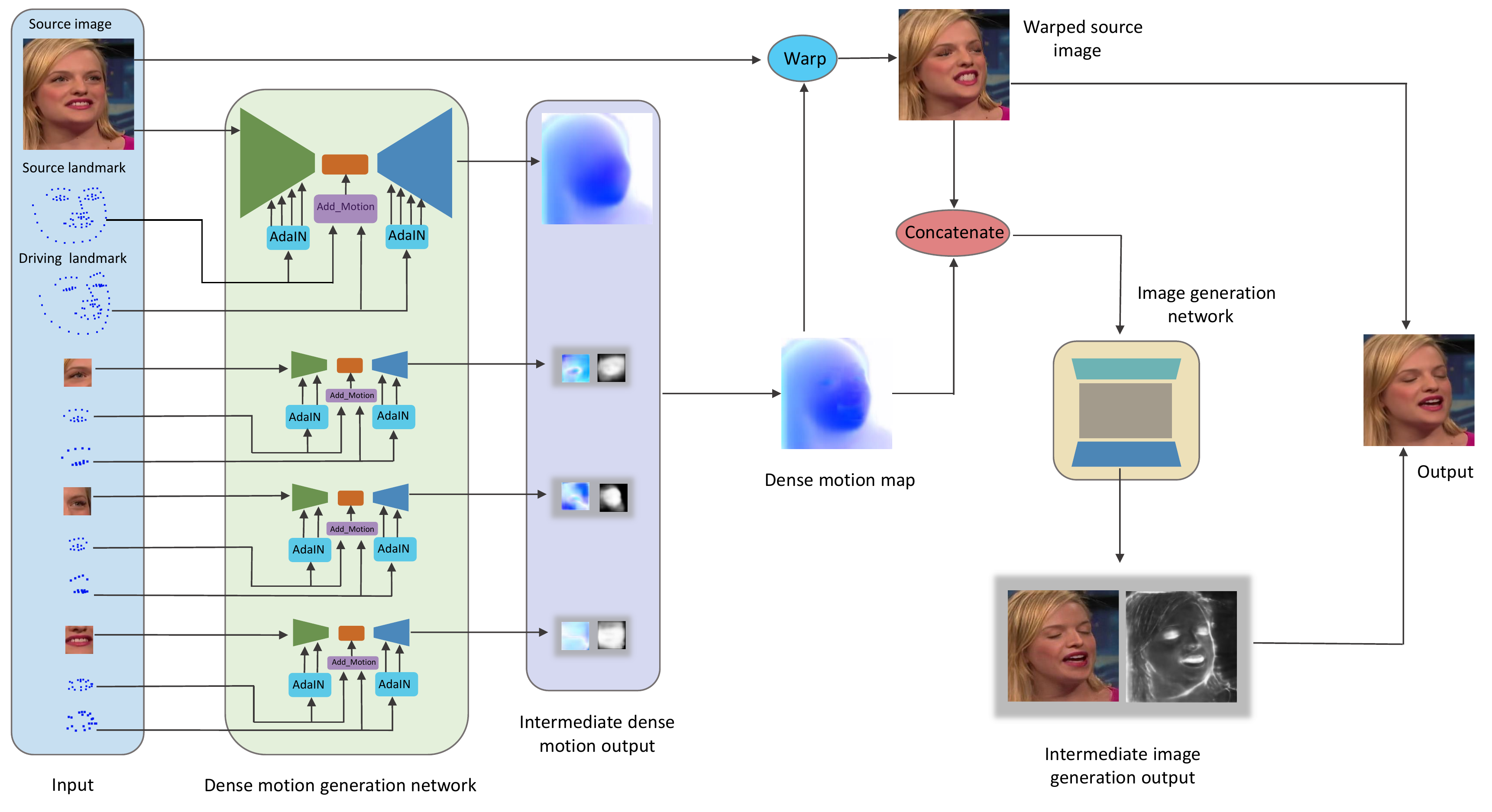}
\end{center}
\caption{Our method consists of a dense motion generation network and an image generation network. The dense motion generation network contains four branches. The global branch works on the full image. The three local branches each focuses on one region. AdaIN and Add\_Motion layers are used for motion transfer in each branch. Outputs of the four branches are combined to get the dense motion map which is used for warping the source image. The warped source image concatenated with the dense motion map is sent to the image generation network to generate a residual image. Warped source image and residual image are weighted and added to get the output image.}
\label{fig:framework}
\end{figure*}
\setlength{\textfloatsep}{7pt}

\section{Related Work}
\subsection{Face Image Animation}
Face image animation has long been studied in computer graphics for its wide application in video games and movies \cite{thies2018headon, thies2016face2face, suwajanakorn2017synthesizing,averbuch2017bringing,kim2018deep,thies2019deferred,vlasic2006face,nagano2018pagan,thies2015real}. However these algorithms often require RGB-D data of a person, have complicated pipelines and generalize poorly. 

Recently several pioneer approaches are developed that do not require personalized training data and can be trained without ground-truth labelling. GANimation \cite{pumarola2018ganimation} builds upon Generative Adversarial Networks (GANs) for the task of facial expression synthesis. It uses a weakly supervised strategy that only requires AU annotations for training. Cycle consistency and attention mechanism are exploited to obtain robust results. X2Face \cite{wiles2018x2face} learns to factorize identity and pose/expression, and generate new face image by warping driving  pose/expression to the source identity. Instead of warping, \cite{zakharov2019few,burkov2020neural,xu2021facecontroller,olszewski2017realistic,song2018talking,tewari2020stylerig} generate new face images in a generative adversarial framework. However these methods have to learn frontalized/neutral face as reference and then add pose/expression to generate new images. 

Several other works choose to decouple motion and appearance \cite{siarohin2019animating, Siarohin_2019_NeurIPS, wang2019few, geng2018warp, gu2020flnet,liu2019liquid,averbuch2017bringing,kim2018deep,ren2020deep}. Monkey-Net \cite{siarohin2019animating} is a framework for general object image animation that consists of a landmark detector, a motion prediction network and an image generation network. The landmark detector can learn to detect landmarks in an unsupervised way. It assumes locally rigid motion model surrounding each landmark. First-order \cite{Siarohin_2019_NeurIPS} assumes locally affine motion model that can model more complex motions. It significantly improves the quality of animated object videos.

\subsection{Driving Modalites}
The driving modality for face image animation can be face landmarks \cite{di2018gp, eskimez2018generating, wang2019few, zakharov2019few, huang2020learning, zakharov2020fast}, text \cite{fried2019text}, audio \cite{thies2020neural, suwajanakorn2017synthesizing, chen2019hierarchical, zhou2020makeittalk, das2020speech, zhou2019talking,chung2017you,vougioukas2019realistic,chen2020talking} and images \cite{Siarohin_2019_NeurIPS, wang2021facevid2vid}. Suwajanakorn et al. \cite{suwajanakorn2017synthesizing} lists a few important practical applications of audio to video generation such as reducing the amount of bandwidth in video transmission, enabling lip-reading from audio for hearing-impaired people and entertainment. Many audio driven face image animation works avoid directly mapping from audio to image, but first convert audio to face landmarks as an intermediate step and then generate images conditioned on synthesized face landmarks \cite{suwajanakorn2017synthesizing, chen2019hierarchical, zhou2020makeittalk, das2020speech}. Zhou et al. \cite{zhou2020makeittalk} points out that face landmarks have low degree of freedom and can help bridge the gap between audio and image by representing the facial characteristics. Chen et al. \cite{chen2019hierarchical} argues that face landmarks can help filter out the noisy signal in audio. Therefore, landmarks driven face image animation is applicable in these audio driven face image animation works. 

\subsection{Landmarks Driven Face Image Animation}
Zakharov et al. \cite{zakharov2019few} learns an identity embedder using a few examples of the source face and combines the rasterized landmarks image and the identity embedded vector in an image generator to synthesize the target image. Their following work \cite{zakharov2020fast} largely improves the speed of neural rendering without compromising the visual quality by considering high frequency details. Wang et al. \cite{wang2019few} generates a new face image by combining optical flow warped version of the input image and the synthesized intermediate image. Optical flow extracted by FlowNet2 \cite{ilg2017flownet} is used as groundtruth for supervision, while we conduct end-to-end learning in an unsupervised manner.

Robust face landmark detection algorithms have been developed over the years. Many existing works \cite{wang2019few, zakharov2019few, zakharov2020fast} use Dlib \cite{dlib09} or FAN4 \cite{bulat2017far} to detect face landmarks for training their algorithms. However, these landmark detectors do not include eye pupils detection. And when applied on face videos, they return results with lots of temporal jittering. We have added eye pupils detection and used methods to stabilize landmark detection on video. Compared with \cite{wang2019few, zakharov2019few, zakharov2020fast}, our method can generate gaze change and our performance is boosted with stabilized face landmarks.

\section{Method}
Given a source face image and a sequence of driving face landmarks, we would like to generate a face video in which the motion of the face resembles the motion of driving face landmarks. Face landmarks are represented using a vector which is the concatenation of coordinates of each landmark. We break down the problem into single face image generation. For each triple of source face image $S_I$, source face landmarks vector $S_{lm}$ and driving landmarks vector $D_{lm}^k$, we generate a new face image $T_I^k$. We then concatenate $T_I^k, k=1,...,n$ sequentially to form a face video, where $n$ is the number of frames.

\subsection{Method Overview}
An overview of our method is shown in Figure \ref{fig:framework}, where we employ a dense motion generation network and an image generation network in an end to end manner for generating a new face image. The dense motion generation network aims to generate a dense motion map $F^k$ that represents the per pixel mapping. We then warp the source image $S_I$ through $F^k$ and get an initial estimation $E_I^k$. Using $f_w^b$ to denote the bi-linear warping operation, we have:
\begin{equation}
E_I^k = f_w^b(S_I,F^k)
\end{equation}

$E_I^k$ and $F^k$ are then concatenated and the result is used as input to the image generation network to synthesize occluded regions and regions that need refinement. With the dense motion map $F^k$ as additional cue, the image generation network better knows where and how $E_I^k$ should be modified and refined. The image generation network outputs a residual image $R_I^k$ and a mask $M^k$. $E_I^k$ and $R_I^k$ and weighted and added to get the final image:
\begin{equation}
T_I^k = E_I^k \odot (1-M^k) + R_I^k \odot M^k
\end{equation}
where $\odot$ is element-wise product. This kind of operation has been employed previously in image animation \cite{grigorev2018coordinate, siarohin2019animating, Siarohin_2019_NeurIPS}, video-to-video synthesis \cite{wang2018video, wang2019few} and image in-painting \cite{yu2018generative} works.

\subsection{Sparse to Dense Motion Transfer}
It is challenging to generate dense motion of an image when only sparse motion on a few landmarks is available. We employ AdaIN layer and Add\_Motion layer as the building blocks to directly conduct motion transfer from landmarks to image. We further combine global and local motion estimation in a unified network to generate not only the global motion but also local fine motion.

\textbf{AdaIN layer for motion transfer.} 
Adaptive Instance Normalization (AdaIN) layer has been invented for image style transfer \cite{karras2019style, huang2017arbitrary,gatys2016image}. In \cite{huang2017arbitrary}, AdaIN performs style transfer by aligning feature statistics of content image to match the style image. The feature statistics are channel-wise mean and variance of feature maps. This shows that the feature statistics can encode style information well. In face image animation literature, AdaIN has been used for transferring face identity from source face image to rasterized landmarks image for new face image generation \cite{zakharov2019few}. 

Different from them, we aim to transfer geometry and motion from sparse landmarks vector to an image. Face landmarks contain rich information about face geometry, including pose, facial expressions and 2D location. By using landmarks vector to control the feature statistics of dense motion generation network through AdaIN, we can make the network carry information of face geometry. Since each feature map is controlled separately with a different mean and variance, different geometry information is transferred to the dense motion network. Our dense motion generation network has an encoder-decoder architecture. 

We employ AdaIN to transfer source face landmark geometry to the encoder and driving face landmark geometry to the decoder. Thus, both source and driving face geometry are encoded in the network. Since our training objective is to reconstruct the target image through warping, the dense motion generation network is optimized to learn the change of face geometry and generate a dense motion map. 

\textbf{Add\_Motion layer for motion transfer.} 
We propose Add\_Motion layer to improve motion transfer from sparse face landmarks to the face image. For Add\_Motion layer, we compute the difference between driving face landmarks vector $D_{lm}^k$ and source face landmarks vector $S_{lm}$ and add it to the hidden representation of the dense motion generation network through a fully connected layer. The Add\_Motion layer directly injects sparse landmark motion to the dense motion generation network. Through the feed-forward decoder, sparse landmarks motion can be converted to dense pixel motion. 

\textbf{Combine global and local dense motion.} 
AdaIN and Add\_Motion layers can transfer satisfying global motion when applied in a dense motion generation network that processes the full image. Without a 3D model, the network can effectively detect the moving foreground and generate face rotation and translation. It can also generate local motion such as eye blinking and mouth opening. However, local motion as subtle as gaze change can not be generated using such one network. Although eyes are only a small part of the face, they can convey important and engaging message. Therefore, it is important to realize gaze change in real practice where we want to make a portrait live. 

To tackle this issue, we propose an approach that combines global and local motion estimation. We have a global branch $M_G$ to estimate the global dense motion map, and three small branches $M_{L_1}$, $M_{L_2}$ and $M_{L_3}$ focus on the left eye eyebrow, the right eye eyebrow and the mouth area, respectively. The input to $M_G$ is $S_I$ of shape $256 \times 256$, $S_{lm}$ and $D_{lm}^k$. The input image to $M_{L_1}$ is a patch of shape $64 \times 64$ cropped around the center of left eye eyebrow, and the input source and driving landmarks vector only contains landmarks that belong to the left eye and eyebrow. We do it in the same way for $M_{L_2}$ and $M_{L_3}$. $M_G$ outputs a dense motion map of size $256 \times 256$. $M_{L_1}$, $M_{L_2}$ and $M_{L_3}$ each outputs a dense motion map and a mask both of size $64 \times 64$. The mask represents how to combine global and local motion map in the corresponding region. It helps making the final motion map smoother at the boundary. The local network is better at capturing subtle motion in its local region. Therefore, we are able to get better dense motion maps and generate more realistic images. $M_{L_1}$, $M_{L_2}$ and $M_{L_3}$ are very small networks that introduce very little computation overhead to the model. The final dense motion map can be denoted as follows:
\begin{equation}
F^k(p) = 
\begin{cases}
f_G^k(p)  & p \notin r_{L_i}^k \\
\\
f_G^k(p) \times (1-m_{L_i}^k(p)) + \\
\hat{f}_{L_i}^k(p) \times m_{L_i}^k(p) &  p \in r_{L_i}^k
\end{cases}
\end{equation}
where $p$ denotes a pixel in the image, $f_G^k$ is the dense motion map output of $M_G$, $r_{L_i}^k, i=1,2,3$ denotes a local region, $m_{L_i}^k$ represents the mask output of $M_{L_i}$. The dense motion map output of $M_{L_i}$ needs to be shifted and scaled before combing with $f_G^k$ and we use  $\hat{f}_{L_i}^k$ to denote the transformed result.

\subsection{Improving Face Landmark Detection}
Improving face landmark detection on videos is crucial for training a better face image animation model. We would like accurate spatial and temporal alignment of face landmarks sequence. Many existing face landmark detectors,  such as Dlib and FAN4 \cite{bulat2017far}, fail to accomplish this goal. We improve face landmark detection on video by combining a heatmap prediction network FAN4 \cite{bulat2017far} with a differentiable regression layer DSNT \cite{nibali2018numerical}. DSNT layer transforms heatmap activation into landmark coordinates. It adds no trainable parameters and is fully differentiable. We additionally augment the regression loss with a registration loss \cite{dong2018supervision, dong2020supervision} for network training on videos. 

Eye pupils convey important information about gaze, therefore we add them in our landmark detection. Our face landmark detector can detect 74 face landmarks.

We initialize our model using pre-trained FAN4 model released by \cite{bulat2017far}. Since our FAN4 model predicts 74 heatmaps rather than 68 in \cite{bulat2017far}, we randomly initialize other weights that correspond to the 6 new heatmaps. We use the WFLW dataset \cite{wayne2018lab} to train our detector. After the training converges, we further fine tune the detector on videos by augmenting the regression loss with the registration loss. 

\subsection{Training Losses}
Our model is trained on a collection of face videos in an unsupervised manner. We perform face landmark detection on the videos before training. We optionally use a state-of-the-art face parsing algorithm \cite{te2020edge} to obtain parsing maps of the videos. During training, we randomly select pairs of source and driving images from each training video. We train the network to reconstruct the driving image $D_I^k$ using the source image $S_I$, source face landmarks vector $S_{lm}$ and driving face landmarks vector $D_{lm}^k$. 

Our training losses consists of the perceptual loss \cite{Siarohin_2019_NeurIPS, johnson2016perceptual}, the least-square GAN loss \cite{mao2017least} and a loss that uses face parsing maps as prior. The perceptual loss is given by:
\begin{equation}
    L_{per}(D_I^k,T_I^k) = \sum_{j=1}^{J}|N_j(D_I^k) - N_j(T_I^k)|
\end{equation}
where $N_j(\cdot)$ is the $j^{th}$ channel feature extracted using VGG-19, $J$ is the number of feature channels in VGG-19. 

The least-square GAN loss consists of two parts. The first part is for training the discriminator $D$ and is given by:
\begin{equation}
    L_{gan}^D(D_I^k,T_I^k) = \lVert 1 - D(D_I^k) \rVert + \lVert D(T_I^k) \rVert
\end{equation}
The second part is for training the generator $G$, given by:
\begin{equation}
    L_{gan}^G(D_I^k, T_I^k) = \lVert 1 - G(T_I^k) \rVert
\end{equation}

We further use face parsing maps as a prior to help guide the training. Given the parsing map for the source image $S_P$, we warp it using the dense motion map $F^k$ and would like the warped result to match the parsing map of the driving image $D_P^k$. The warping operation we use is the nearest warping denoted as $f_w^n$. We compare $f_w^n(S_P)$ with $D_P^k$ and the objective is to minimize the percentage of pixels where the parsing labels are not matched. The loss function is given by:
\begin{equation}
   L_{par}(D_I^k, T_I^k) = 1-|f_{eq}(f_w^n(S_P,F^k),D_P^k)|
\end{equation}
where $f_{eq}$ is a logic operation function that conducts element-wise comparison of two matrices. It returns 1 if two elements are equal, and 0 otherwise. 

The final objective is to minimize the total of all the above losses:
\begin{equation}
L_{total}=L_{per}+L_1+L_{gan}+L_{par}
\end{equation}

\begin{figure*}
\begin{center}
\includegraphics[width=1.0\textwidth]{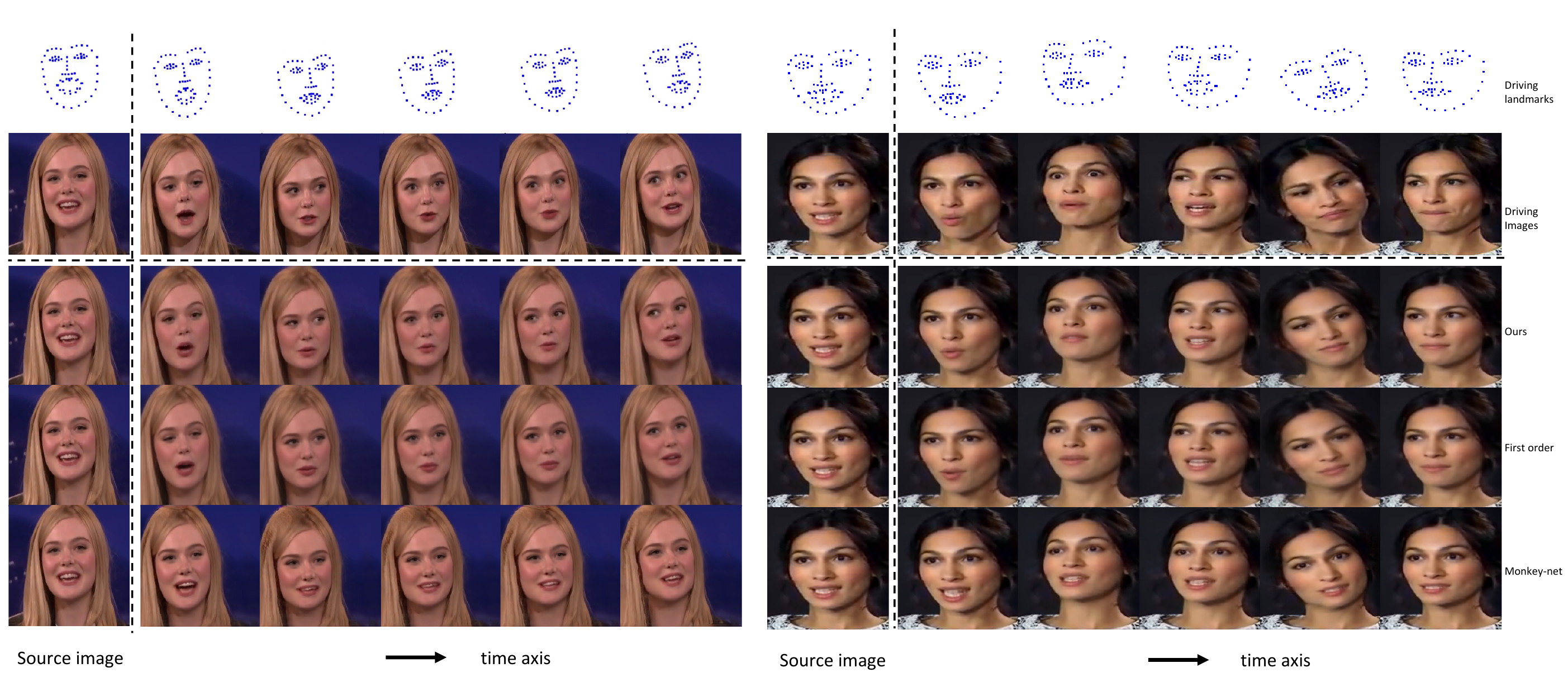}
\end{center}
\caption{Qualitative results of same identity testing on VoxCeleb1. Our method can generate more accurate gaze change.}
\label{fig:same_vox}
\end{figure*}
\setlength{\textfloatsep}{7pt}

\section{Experiments}
We conduct experiments on two datasets, VoxCeleb1 \cite{Nagrani19, Chung18b,Nagrani17} and FaceForensics \cite{roessler2019faceforensicspp}. 

VoxCeleb1 is a face video dataset with 22496 videos collected from YouTube. We follow \cite{Siarohin_2019_NeurIPS} for image pre-processing. During pre-processing, each image is cropped and resized to $256\times 256$. We obtain 12322 videos for training and 512 videos for testing. We use our face landmark detector to extract face landmarks from the videos. 

FaceForensics \cite{roessler2019faceforensicspp} is a much smaller dataset that contains 1004 face videos acquired from YouTube. As in \cite{gu2020flnet} we randomly split the videos into 75\% for training and 25\% for testing. The videos in FaceForensics are pre-processed in the same fashion as VoxCeleb1. 

\begin{table*}
  \caption{Comparison with state-of-the-arts on VoxCeleb1 dataset.}
  \label{tab:sameidentity}
  \centering
  \begin{tabular}{l|l|l|l|l|l|l|l}
    \hline
    Method & $L_1$ $\downarrow$ & FID $\downarrow$ & SSIM $\uparrow$ & LPIPS $\downarrow$ & MS-SSIM $\uparrow$ & AKD $\downarrow$ & PSNR $\downarrow$            \\
    \hline
    Few-shot(2019) \cite{zakharov2019few}   &  N/A & 30.6 & 0.7200  & 0.75 & N/A & N/A & N/A\\
    PuppeteerGAN (2020) \cite{chen2020puppeteergan} & N/A & 33.61 & 0.7255  & N/A & N/A & N/A & N/A \\
    Monkey-Net(2019)  \cite{siarohin2019animating} & 0.0490 & 12.66 & 0.7367 & 0.1249 & 0.7966 & 1.76 & 30.86 \\
    First-order(2019) \cite{Siarohin_2019_NeurIPS} & \textbf{0.0415} & 10.69 & \textbf{0.7936} & 0.1020 & \textbf{0.8587} & 1.09 & \textbf{30.97} \\
    Ours              & 0.0437    &  \textbf{9.06} & 0.7770 & \textbf{0.1004} & 0.8438 & \textbf{0.89} & 31.06 \\
    \hline
  \end{tabular}
\end{table*}

\subsection{Evaluation Metrics}
The following metrics \cite{Siarohin_2019_NeurIPS, gu2020flnet, wang2021facevid2vid, zakharov2020fast} are employed in our experiments to quantitatively evaluate our method. 

$L_1$: The average $L_1$ distance between two images.

PSNR: The mean squared error between two images.

FID score \cite{heusel2017gans}: It measures perceptual realism by extracting feature embeddings using Inception network and computing the average euclidean distance.

SSIM/MS-SSIM \cite{wang2004image}: They compare similarity between two images based on luminance, contrast and structure. MS-SSIM is a multi-scale variant of SSIM.

LPIPS \cite{zhang2018perceptual}: It measures the perceptual similarity between two images by computing cosine distance of each channel and averaging across channels of the conv1-conv5 layers of AlexNet.

AKD: We do face landmark detection using our face landmark detector and compute the average pixel distance. It measures how well the motion is preserved.

\subsection{Experimental results}
\subsubsection{Same identity testing results}

We first conduct the same identity testing on VoxCeleb1 and FaceForensics. 
For VoxCeleb1, we use the same quantitative evaluation setting as \cite{Siarohin_2019_NeurIPS}. For each testing video, the first frame is used as the source image and all of the frames in the video are used as driving images. We compare our results with the state-of-the-arts, First-order \cite{Siarohin_2019_NeurIPS}, Monkey-Net \cite{siarohin2019animating}, Few-shot \cite{zakharov2019few} and PuppeteerGAN \cite{chen2020puppeteergan}, X2Face \cite{wiles2018x2face} in Table \ref{tab:sameidentity}.  We use sparse face landmarks for animating the face image, same for Few-shot and PuppeteerGan, while the other methods use images for driving. Our method performs comparable with First-order and better than all the other methods. Some qualitative results are shown in Figure \ref{fig:same_vox}. Our method can generate images with subtle local motion. Using the two landmarks on the pupils, we are able to better control the gaze than First-order while other methods have difficult doing this. 

For FaceForensics, we randomly select one image as source image and use all the other images as driving images for each video. We report quantitative comparison results in Table \ref{tab:FaceForensics}. Note that in \cite{gu2020flnet, wiles2018x2face}, 16 source images are used while we only use one source image, however, we still achieve better results than them.  

\begin{table}
  \caption{Comparison of cross identity testing on VoxCeleb1.}
  \label{tab:crossidentity}
  \centering
  \begin{tabular}{l|l|l}
    \hline
    Method & FID $\downarrow$ & User preference $\uparrow$            \\
    \hline
    Monkey-net(2019) \cite{siarohin2019animating} & 70.7 & \hskip 7mm 12.2\%  \\
    First-Order(2019) \cite{Siarohin_2019_NeurIPS} & 57.43 & \hskip 7mm 37.8\% \\
    Ours     &    \textbf{55.81} & \hskip 7mm \textbf{49.9}\% \\
    \hline
  \end{tabular}
\end{table}

\begin{table}
  \caption{Comparison with state-of-the-arts on FaceForensics.}
  \label{tab:FaceForensics}
  \centering
  \begin{tabular}{l|l|l}
    \hline
    Method & $L_1$ $\downarrow$ & FID $\downarrow$                \\
    \hline
    GANimation(2018) \cite{pumarola2018ganimation}   & 16.19 & 47.99  \\
    X2Face(2018) \cite{wiles2018x2face}  & 11.05 & 23.98 \\
    FLNet(2020) \cite{gu2020flnet} & 10.20 & 20.62 \\
    First-Order(2019) \cite{Siarohin_2019_NeurIPS} & 14.06 & 13.99 \\
    Ours     &    \textbf{10.05}  & \textbf{10.33}\\
    \hline
  \end{tabular}
\end{table}

\subsubsection{Cross identity testing results}
We then conduct cross identity testing. We report quantitative and qualitative cross identity testing results on VoxCeleb1. To do quantitative evaluation on VoxCeleb1, we first select frontal face images in the testing videos. We obtain the 3D landmarks of each face using the  3D face landmark detector \cite{newell2016stacked}. We then get the 3D face pose by aligning the 3D landmarks to a template. We further select images with pose less than $5^{\circ}$ for pitch yaw and roll. For each testing video, we get a clip that starts with its selected frontal face image. We select pairs of clips whose starting images are very close in face poses. We use $1^{\circ}$ as the threshold. 702 pairs of video clips are obtained for testing. 

We compute FID score and compare with First-order \cite{Siarohin_2019_NeurIPS} and Monkey-Net \cite{siarohin2019animating}. The results are shown in Table \ref{tab:crossidentity}. We achieve a better FID score than them. We also conduct a user study to compare our results with First-order and Monkey-net. In the user study, we randomly select 100 out of the 702 examples in cross identity testing. We ask 18 participants to select the best quality video among the three methods for each example. The results are shown in Table \ref{tab:crossidentity}. Our results are most preferred among the three methods. 

\begin{figure}
\begin{center}
\includegraphics[width=0.48\textwidth]{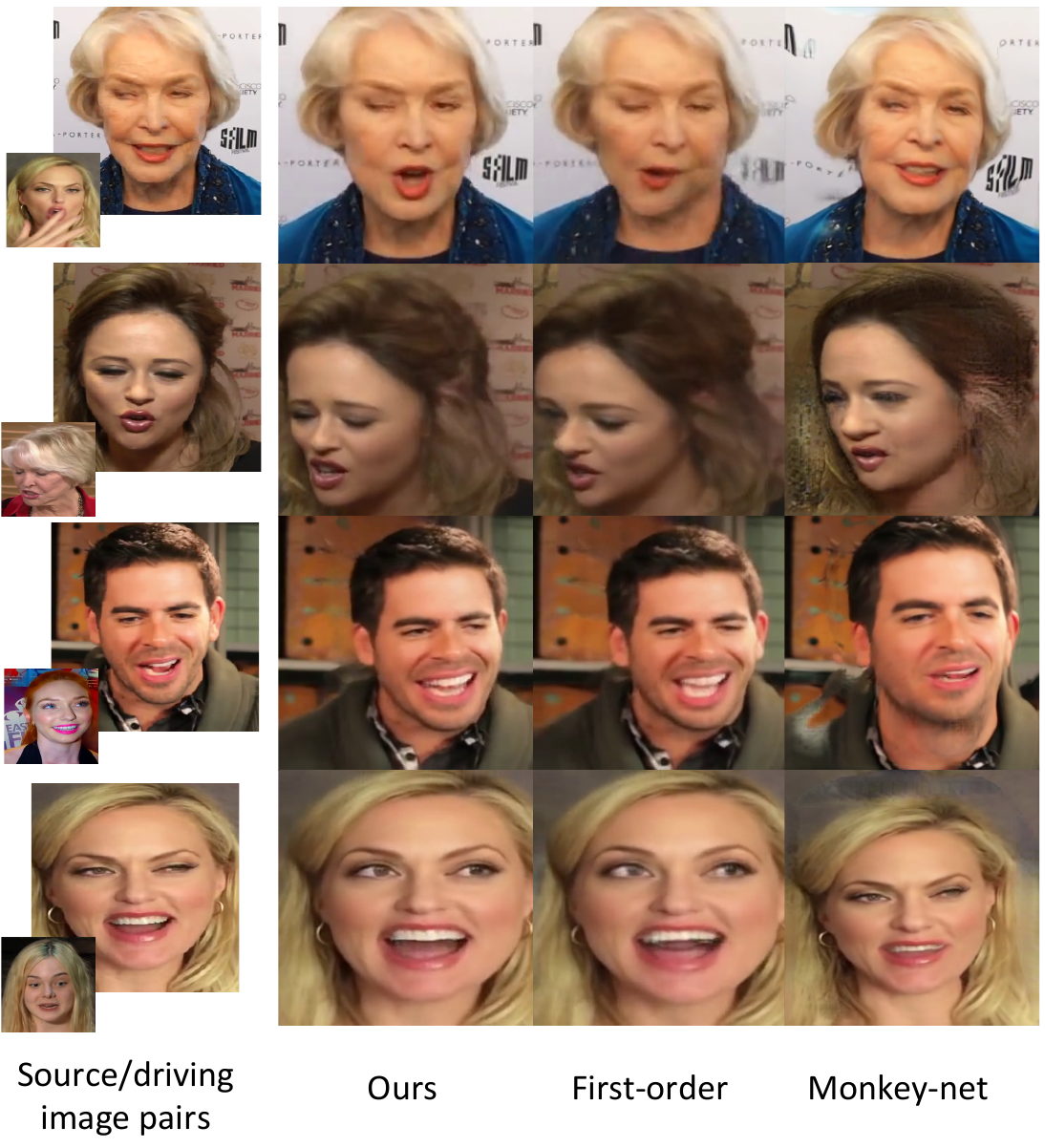}
\end{center}
\caption{Qualitative results of cross identity testing on VoxCeleb1. Our method is more robust to hairstyle change and occlusion caused by hands. We can also generate images with higher perceptual quality.}
\label{fig:cross_vox}
\end{figure}

\begin{figure}
\begin{center}
\includegraphics[width=0.48\textwidth]{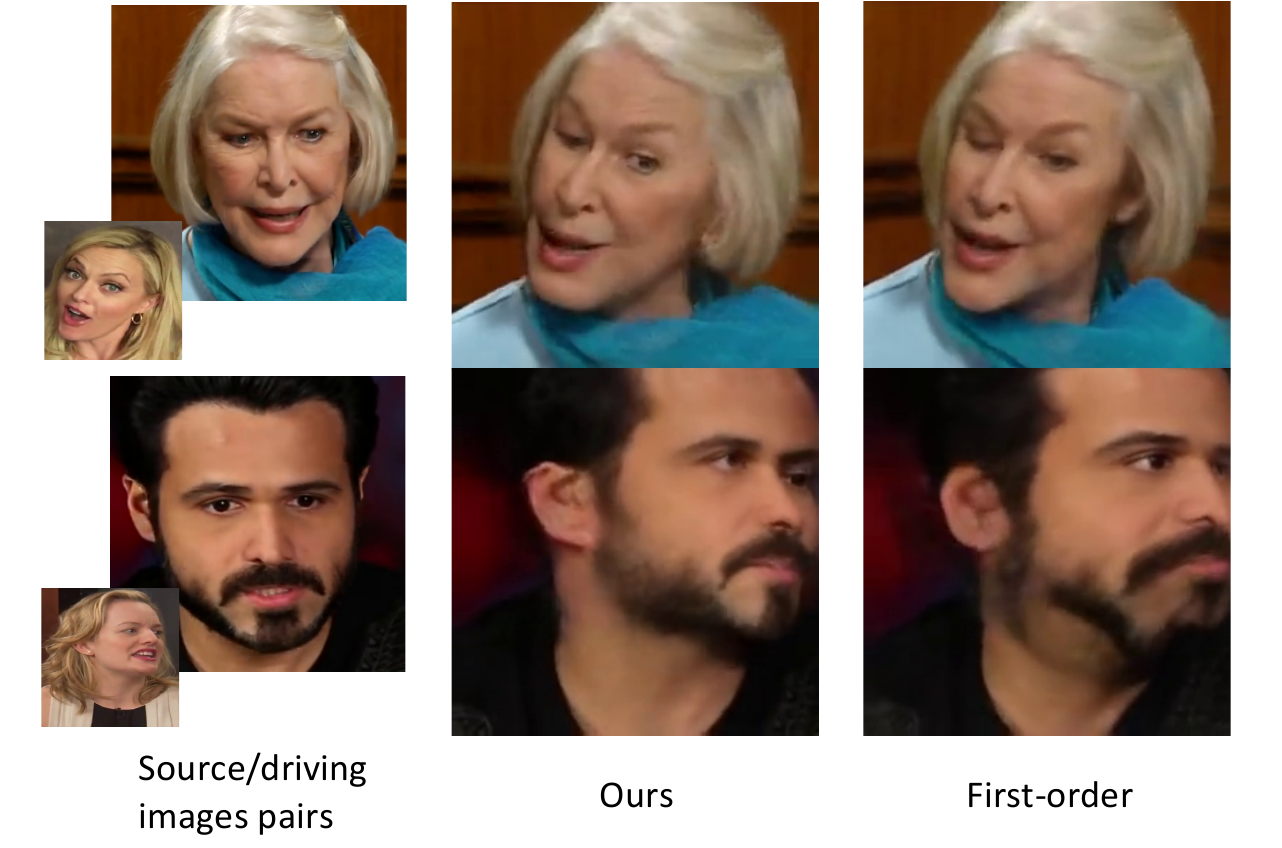}
\end{center}
\caption{Some fail cases. When the face has large pose change or the face is close to the image boundary, both our method and First-order produce artifacts.}
\label{fig:fail_cases}
\end{figure}

Some qualitative results are shown in Figure \ref{fig:cross_vox}. When the hairstyles between source face and driving face are very different or when there are occlusions caused by hands, First-order produces more distortion than ours. In Figure \ref{fig:fail_cases} we show some failing cases. When there is large pose change or when the face is close to the image boundary, both our method and First-order fail in the eye area. 
\vspace{-5pt}

\subsubsection{Cross identity testing of images in the wild} 
\begin{figure}
\begin{center}
\includegraphics[width=0.48\textwidth]{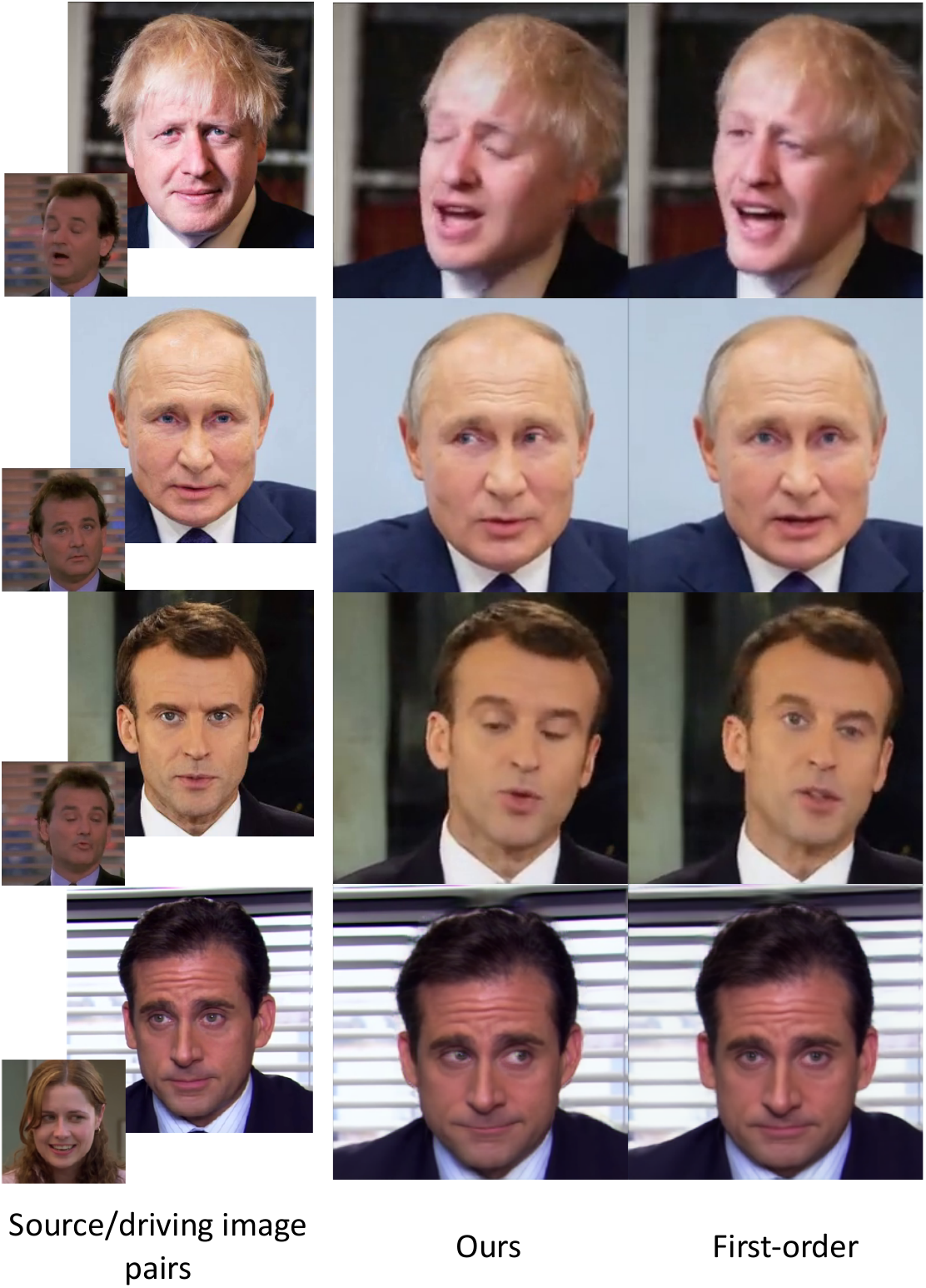}
\end{center}
\caption{Images in the wild testing. Our method can generate faces with more flexible eyes motion and less shape distortion than First-order.}
\label{fig:office}
\end{figure}

We further perform the challenging images in the wild testing where we download leaders and celebrity images from the Internet and use our model trained on VoxCeleb1 to generate face images. The testing is also across different identities and we use face image of one person to drive the motion of face image of another person. We compare with First-order and its model is also trained on VoxCeleb1. We show visualization results in Figure \ref{fig:office}. From Figure  \ref{fig:office}, images generated by our method look more realistic with vivid gaze change and have less shape distortion. While images generated by First-order can not follow the direction of the eyes in the driving images. 

\setlength{\textfloatsep}{7pt}

\setlength{\textfloatsep}{7pt}

\begin{figure}
\begin{center}
\includegraphics[width=0.48\textwidth]{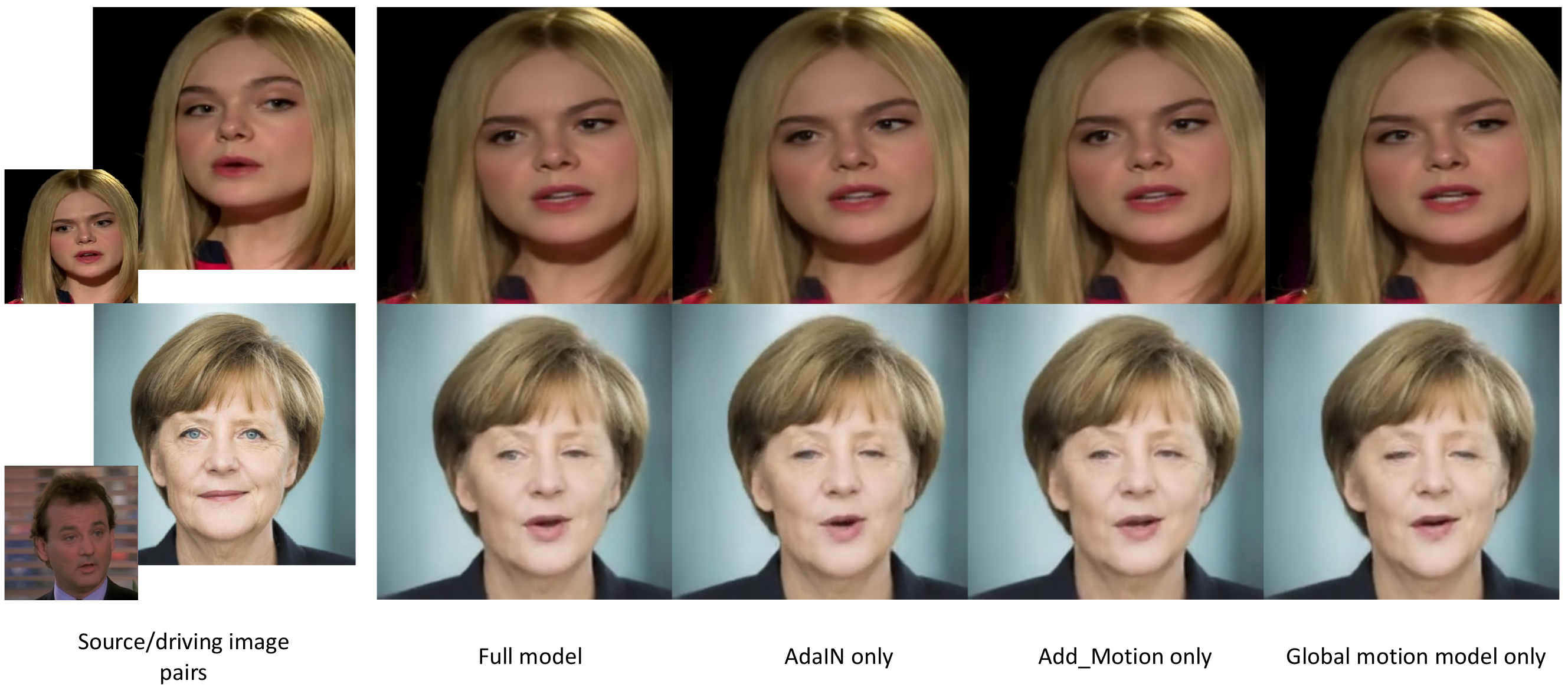}
\end{center}
\caption{Examples of ablation study on motion transfer. The first example shows that the global motion only model can not generate eye movement. The second example shows that image generated by the full model has higher quality in the eyes and mouth region.}
\label{fig:motionablation}
\end{figure}
\setlength{\textfloatsep}{7pt}

\setlength{\textfloatsep}{7pt}

\subsection{Ablation Study}
\subsubsection{Effectiveness of motion transfer strategies}
We conduct ablation study to show the effectiveness of AdaIN layer, Add\_Motion layer  and the global \& local approach for motion transfer. We consider the following three settings. In the first setting, we remove AdaIN layers in the dense motion generation network. In the second setting, we remove Add\_Motion layers in the dense motion generation network. And in the third setting, we remove the three local branches in the dense motion generation network. We train and test the three models on VoxCeleb1.

The quantitative comparison results are shown in Table \ref{tab:ablation}. One can see that the full model achieves the best results, showing that all the components contribute to motion transfer in the dense motion generation network. We provide qualitative comparison in Figure \ref{fig:motionablation}. Compared with AdaIN only and Add\_Motion only model, the full model generates images with higher quality, especially in the eyes and mouth region. Without the local motion model, the global motion only model can not generate eye movement.

Note in this ablation study, we only use warped source image as input to the image generation network without concatenating it with the dense motion map. Comparing our results in Table \ref{tab:sameidentity} and Table \ref{tab:ablation}, we can see that adding dense motion map as additional cue to the image generation network improves the peformance.

\begin{table}
  \caption{Ablation study results on motion transfer.}
  \label{tab:ablation}
  \centering
  \begin{tabular}{l|l|l|l}
    \hline
    Setting & $L_1$ $\downarrow$ & FID $\downarrow$ & SSIM $\uparrow$             \\
    \hline
    Full model  & \textbf{0.0442} & \textbf{9.82} & \textbf{0.7733} \\
    no Add\_Motion layer & 0.0449 & 10.20 & 0.7728 \\
    no AdaIN layer & 0.0449 & 9.98 & 0.7725\\
    no local motion & 0.0454 & 10.17 & 0.7729 \\
    \hline
  \end{tabular}
\end{table}

\subsubsection{Face landmark detector improvement}
To show the improvement of our face landmark detector, We compare the four face landmark detectors, Dlib, FAN4, FAN4+DSNT and FAN4+DSNT+SBR. The experiment is conducted on VoxCeleb1 dataset.

We first study their face landmark detection performance on video. To measure temporal smoothness of landmark trajectories, we compute the following metric:
\begin{equation}
    SMS = \frac{1}{T}\sum_{t=1}^{T}|(D_{lm}^{t-1} + D_{lm}^{t+1}) / 2 - D_{lm}^t|
\end{equation}
where $D_{lm}^{t-1}$, $ D_{lm}^t$, $D_{lm}^{t+1}$ are three consecutive landmark frames, $T$ is the number of frames. 
We also conduct a user study where we give 10 participants 10 groups of face videos with landmarks marked. We ask the participants to select the video that has the most stable and accurate face landmarks in each group. The videos we use are randomly selected from VoxCeleb1. The results are shown in Table \ref{tab:facedetector}. It shows that FAN4+DSNT+SBR is most favored for face landmark detection. 

We then train our face image animation model using face landmarks detected by each of the detectors. We use 68 landmarks in this experiment since Dlib and FAN4 only detect 68 landmarks. We only use perceptual loss for training. The performance comparison is shown in Table \ref{tab:facedetector}. Using FAN4+DSNT+SBR as the face landmark detector, we are able to boost face image animation performance.


\begin{table}\small
  \caption{Ablation study results on face landmark detector.}
  \label{tab:facedetector}
  \centering
  \scalebox{0.93}{
  \begin{tabular}{l|l|l|l|l|l}
    \hline
    Detector & \begin{tabular}{@{}c@{}}User \\ Study $\uparrow$ \end{tabular}  & SMS $\downarrow$ & $L_1$ $\downarrow$ & FID $\downarrow$ & SSIM $\uparrow$              \\
    \hline
    Dlib   & 0 \% & 92 & 0.0479 & 12.98 & 0.7640 \\
    FAN4   & 0 \% & 115 & 0.0478 & 12.71 & 0.7662 \\
    FAN4+DSNT   & 38 \% & 61  & 0.0460 & 11.88 & 0.7759 \\
    \begin{tabular}{@{}c@{}}FAN4+DSNT \\ SBR \end{tabular}    & \textbf{62\%} & \textbf{51} & \textbf{0.0457} & \textbf{11.70} & \textbf{0.7775}\\
    \hline
   \end{tabular}}
\end{table}

\subsection{Limitations}
Although we have taken effective ways to improve the quality of generated images, they still look blurry especially when there is large pose change. State-of-the-art super resolution methods \cite{shi2016real, xiang2021zooming} may be used to improve the resolution of generated images. Our image generation network has the same architecture as the one in \cite{Siarohin_2019_NeurIPS}. Techniques from ProGAN \cite{karras2017progressive} and StyleGAN \cite{karras2019style} may be borrowed to improve the image generation performance. In addition, the face landmarks we use are very sparse. Some detailed facial motion can not be captured. Denser landmarks could be used to provide more motion guidance. Other modalities such as audio may be used for further improvement.

\section{Conclusion}
A novel method for face image animation driven by sparse face landmarks has been presented. We adopt the dense motion generation followed by image refinement approach. We propose to use AdaIN layer and our Add\_Motion layer for transferring motion from sparse face landmarks to the face image. By combining global and local motion generation, our method can generate not only global motion, such as face rotation and translation, but also fine local motion, such as gaze change. We further improve face landmark detection on video. It greatly improves the visual quality and temporal coherence of generated videos. Experiments have shown that our method achieves comparable results to the state-of-the-art image driven methods on same identity testing and better results on cross identity testing. In the future, we are interested in using the audio modality to further improve the quality of generated face videos.


\section{Acknowledgements}
Thank Jia Wang for helping downloading and transmitting the VoxCeleb1 dataset. Thank our Reviewer 3 for suggesting adding dense motion map to the image generation network that improves our method.


\end{document}